\documentclass[runningheads]{llncs}
\usepackage{xcolor}

\usepackage{graphicx}
\usepackage{listings}
\usepackage{geometry}
\begin{document}
\title{DPCL: a Language Template for Normative Specifications}
%
%
\author{Giovanni Sileno$^1$, Thomas van Binsbergen$^1$, Matteo Pascucci$^2$, Tom van Engers$^1$}
\authorrunning{G. Sileno et al.}
%
\institute{$^1$Informatics Institute, University of Amsterdam, the Netherlands\\
$^2$Slovak Academy of Sciences, Bratislava, Slovakia\\
\email{g.sileno@uva.nl}}
\maketitle              
\lstset{basicstyle=\ttfamily\footnotesize}

\vspace{-20pt}

\section{Introduction}

Several solutions for specifying normative artefacts (norms, contracts, policies) in a computational processable way have been presented in the literature. 
Amongst the most recent efforts we acknowledge e.g.  LegalRuleML \cite{Palmirani2011,Lam2019}, PROLEG \cite{Satoh2011}, InstAL \cite{Padget2016}, ODRL \cite{Iannella2018,DeVos2019}, Symboleo \cite{Sharifi2020}, FLINT/eFLINT \cite{VanDoesburg2016,VanBinsbergen2020}, and Logical English \cite{Kowalski2021}. Legal core ontologies (e.g. LKIF-core \cite{Hoekstra2007}, UFO-L \cite{Griffo2018}) have been proposed to systematize concepts and relationships relevant to normative reasoning. However, no solution amongst those has achieved general acceptance. Even more unexpectedly, there exists no common ground (nor representational, nor computational) enabling us to compare these solutions. 
Yet, all these efforts share the same motivation (representing normative directives), therefore it is plausible that there may be a representational model encompassing all of them. 

This presentation will introduce DPCL, a domain-specific language (DSL) for specifying higher-level policies (including norms,  contracts, etc.) as computational artefacts,  centred on Hohfeld's framework of fundamental legal concepts (\textbf{D}uty, \textbf{P}ower, \textbf{C}laim, \textbf{L}iability, going together with their opposites Liberty, Disability, No-Claim, Immunity) \cite{Hohfeld1913,Hohfeld1917}, echoing in turn Salmond's work \cite{Salmond1907}. Contrary to the representation formalisms in the list presented above, but similarly to ODRL, DPCL has to be seen primarily as a “template”, i.e. as an informational model for architectural reference, rather than a fully-fledged formal language. 
DPCL aims to make  explicit the general requirements that should be expected in a language for norm specification. In this respect, it goes in the direction of legal core ontologies, or more in general, of legal knowledge interchange formats. However, our proposal aims to keep the character of a DSL, rather than a set of axioms in a logical framework: 
it is meant to be cross-compiled to underlying languages/tools adequate to the type of target application (e.g. BGP policies for routing services \cite{Sileno2021}).
In terms of features, the DPCL language is meant to promote three fundamental innovations in a unified framework: 
\begin{enumerate}
    \item it brings \textit{power} constructs to the foreground, i.e. mechanisms of \textit{institutional change}, as necessary elements for the operationalization of norms, even if they are not required in absolute sense to specify norms (cf. deontic models); they provide the ``chain of transmission'' required to pass from deontic specifications to models of norms compiled out operationally. 
    \item it highlights the necessity of considering by design an ``ecology'' of normative artefacts, by enabling the embedding of a \textit{plurality of references} (e.g., sources of law) to be taken into account for normative reasoning, and by specifying mechanisms for \textit{resolving potential conflicts}.
    \item it opens up the possibility of an alternative semantics with respect to traditional extensional/set-based semantics, tailored to capture the specificity of \textit{qualification} in law.
\end{enumerate}
This document will provide an overview on some of the language design choices we are currently working on.\footnote{The present work is part of a broader research effort. Previous works extensively investigated the concept of power, proposing semantics based on Event Calculus \cite{Sileno2019}, modal logic \cite{Sileno2020}, and benchmarking computational efficiency in integrating causal and logical dependencies \cite{Sileno2020b}. Our work on 
normative relationships resulted in the identification of different abstractions of power (force-centred, outcome-centred, and change-centred) \cite{Pascucci2021}, and in the development of FLINT/eFLINT \cite{VanDoesburg2016,VanBinsbergen2020}, a frame-based DSL.}

\section{Language fundamentals}

\paragraph{Objects and Events} At a fundamental level, DPCL relies on the common-sensical distinction of objects (-types) from events (-types), an ontological ground confirmed e.g. in LKIF-core \cite{Hoekstra2007}. 
All objects (including states) are denoted as literals, as e.g. \texttt{user}. Transition events are instead denoted with prefixed literals as e.g. \texttt{\#borrow} or \texttt{\#lend}. When transient, events should rather be modeled as objects, as e.g. \texttt{borrowing}, or \texttt{raining}, to denote for their ongoing state. A special case of transition events are those concerning the \textit{creation} or \textit{removal} of objects. For instance, the activation of a \texttt{raining} state is denoted with \texttt{+raining}, whereas its disactivation as \texttt{-raining}.

\paragraph{Parameters, refinements} Objects (and events) can have internal properties that can be specified or refined via the program. 
For instance, an action (or action-type) could be specified further referring to the thematic roles of verbs used in linguistics (\texttt{agent}, \texttt{patient}, \texttt{recipient}, \texttt{instrument}, etc.). For the description of internal components, DPCL follows a JSON-like syntax. Possible internal properties may be accessed as in OOP, e.g. \texttt{user.online}. To reify the informational model, certain objects (e.g. the normative positions) come with fixed parameters. 

\paragraph{Transformational vs Reactive}
DCPL considers the distinction of transformational and reactive rules, proceeding along LPS \cite{Kowalski2012}. Transformational rules follow the template: \texttt{a -> b.}, meaning that if \texttt{a} holds, then \texttt{b} holds too. Interpreting the ``if'' as ``as long as'', the transformation aspect (of \texttt{a} into \texttt{b}) becomes explicit. Reactive rules in DCPL are in the form: \texttt{\#f => \#g.}, meaning that the occurrence of a \texttt{\#f} event triggers subsequently a \texttt{\#g} event. Causal effects become explicit with production events, e.g. \texttt{\#rain => +wet}.

\paragraph{Descriptors}
Extensional semantics applied in formal languages are based on set-theory and consider classes or types as sets of instances. This means that they (theoretically) maintain access to an abstract data structure, used e.g. to enlist all instances, and then to check whether an instance belongs to it. 
Rather than sets of objects centrally maintained, DPCL considers the association of several descriptors to each object. 
As an analogy, consider how people generally keep in their own wallets id-cards, library cards, discount coupons, membership cards, etc. These are examples of several “identity” certificates (unique or not) operational in different contexts, which have been issued by an authority, and of which no other copy may exist, except the one people keep with them. 
DPCL aims to bring the \textit{qualification} act in the foreground, and it does so by means of the \texttt{in} operator, as e.g. \texttt{holder in member} which provides a \texttt{holder} object  with a \texttt{member} descriptor. 

\paragraph{Normative positions}
Hohfeld's framework of normative concepts includes 8 notions, distributed on 2 squares, concerning respectively deontic (or duty-related) and potestative (or power-related) directives. Similarly to eFLINT \cite{VanBinsbergen2020}, these concepts corresponds in DPCL to  object-types with certain fixed parameters (\textit{frames}, or stereotypical knowledge constructs). Their instantiation, as that of any other object, can be conditioned by transformational or reactive rules.



\paragraph{Composite objects}
More complex institutional concepts (e.g. ownership) can be described as a composition of primitive ones. DPCL allows for combining the objects above into reusable, parameterizable compounds. Note that if there is nothing else than duty and power frames in those compounds, the institutional/non-institutional knowledge decomposition is evident. We could in principle use any other means (ontologies, logic programs, etc.) for the domain and/or terminological knowledge.

\section{Example}

Library regulation is a classic example used in the normative systems literature, and we will use it here as well to illustrate the main features of the language. 
\begin{itemize}
    \item \textit{Student or staff can register as member of the library by using their id card} (example of act creating a new qualification).\vspace{-8pt}
\end{itemize}
\begin{lstlisting}
    power {
        holder: student | staff
        action: #register { instrument: holder.id_card }
        consequence: holder in member
    }
\end{lstlisting}
\begin{itemize}\vspace{-5pt}
    \item \textit{Any member can borrow a book for at max 1 month} (act creating a composite object). \vspace{-6pt}
\end{itemize}
\begin{lstlisting}
    power {
        holder: member
        action: #borrow { item: book }
        consequence: +borrowing {
            lender: library
            borrower: holder
            item: book
            timeout: now() + 1m 
        }
    }
\end{lstlisting}
\begin{itemize}\vspace{-4pt}
    \item \textit{By borrowing, the borrower can be requested in any moment to return the item. The borrower has the duty to return the item within the given date. If the borrower does not return the item, (s)he may be fined} (example of composite object, with power, duty and violation constructs). \vspace{-3pt}
\end{itemize}
\begin{lstlisting}
    borrowing(lender, borrower, item, timeout) {
        power {
            holder: lender
            action: #request_return
            consequence: +duty {
                holder: borrower
                counterparty: lender
                action: #return { item: book }
            }
        }
        duty d1 {
            holder: borrower
            counterparty: lender
            action: #return { item: book }
            violation: now() > timeout
        }
        +d1.violation => +power {
            holder: lender
            action: #fine
            consequence: +fine(borrower, lender)
        }
    }
\end{lstlisting}\vspace{2pt}

The violation construct is particularly interesting as it is a common feature in deontic-based languages (e.g. LegalRuleML \cite{Palmirani2011}). However, when operationalizing a norm, there always needs  to be some actor who monitors and declares the occurrence of such violation. DPCL has sufficient granularity to capture this. We can for instance rewrite the \texttt{violation: now() > timeout} specification in \texttt{duty d1} in terms of power associated to the claimant (counterparty of the duty-holder):
\begin{lstlisting}
        now() > timeout -> power {
            holder: d1.counterparty
            action: #declare_violation { target: d1 }
            consequence: +d1.violation 
        }
\end{lstlisting}
These and other transformations (used e.g. for unveiling general normative patterns) can be easily generalized by means of rewriting rules. 

\section{Perspectives}
DPCL aims to integrate the accessibility of ODRL \cite{Iannella2018}, with ontological intuitions expressed in LKIF-core \cite{Hoekstra2007} and in LPS \cite{Kowalski2012} (at the base of Logical English \cite{Kowalski2021}), and the finer representational granularity of Hohfeld's framework, here expressed in a frame-based form as in FLINT/eFLINT \cite{VanDoesburg2016,VanBinsbergen2020}. Our hypothesis is that adequate rewriting rules can be used to transform (or compile out) deontic specifications into potestative ones, and into operational code. Via the inverse problem, we may infer ``revealed'' deontic preferences from e.g. potestative specifications. If this is true, DPCL has the potential of becoming a \textit{lingua franca} informational model for normative languages. Our future work aims to test this hypothesis.  

\bibliographystyle{splncs04}
\bibliography{biblio}

\end{document}